\documentclass{article}

\usepackage[table]{xcolor}
\usepackage[final]{corl_2026} 
\makeatletter
\renewcommand{\@conferencelocation}{Austin, Texas, USA}
\makeatother

\usepackage{amsmath}
\usepackage{amssymb}
\usepackage{amsfonts}
\usepackage{graphicx}
\usepackage{caption}
\usepackage{booktabs}
\usepackage{multirow}
\usepackage{float}
\usepackage[ruled,vlined,linesnumbered]{algorithm2e}
\DontPrintSemicolon
\SetAlFnt{\footnotesize}
\SetAlCapFnt{\small}
\SetAlCapNameFnt{\small\bfseries}
\SetAlgoNlRelativeSize{-1}
\SetNlSty{textbf}{}{}
\SetKw{Return}{return}

\title{Hiking in the Wild: A Scalable Perceptive Legged Parkour Framework for Humanoid Robots}

\author{%
Shaoting Zhu$^{1,2,*}$ \enspace Ziwen Zhuang$^{1,2,*}$ \enspace Mengjie Zhao$^{1}$ \enspace Kun-Ying Lee$^{3}$ \enspace Hang Zhao$^{1,2,\dagger}$\\
$^{1}$IIIS, Tsinghua University \quad $^{2}$Shanghai Qi Zhi Institute\\
$^{3}$Department of Computer Science and Technology, Tsinghua University\\
$^{*}$Equal contribution \quad $^{\dagger}$Corresponding author
}

\hypersetup{
  pdftitle={Hiking in the Wild: A Scalable Perceptive Legged Parkour Framework for Humanoid Robots},
  pdfauthor={Shaoting Zhu, Ziwen Zhuang, Mengjie Zhao, Kun-Ying Lee, Hang Zhao}
}

\begin{document}
\maketitle

\begin{figure}[H]
  \centering
  \includegraphics[width=\linewidth]{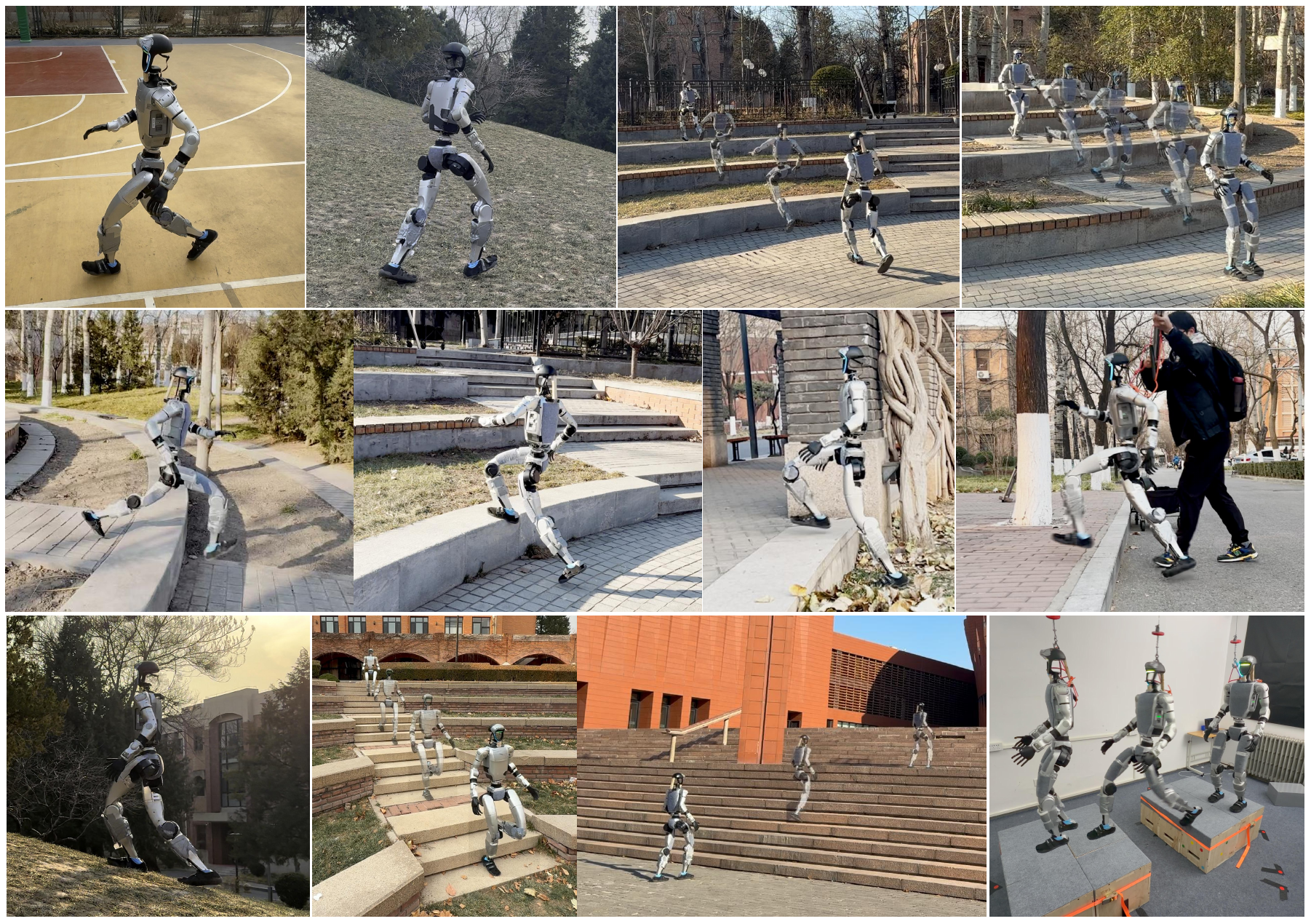}
  \caption{\textbf{Hiking in the Wild.} Our framework enables a humanoid robot to traverse diverse terrains in the wild. The robot can run at a maximum speed of 2.5 m/s over complex terrain. It can also negotiate stairs, gaps, high platforms, and ramps. Relying on depth images for perception, our extensible end-to-end framework achieves robust performance across these scenarios.}
  \label{fig:teaser}
\end{figure}

\begin{abstract}
Achieving robust humanoid locomotion in complex, unstructured environments requires transitioning from reactive proprioception to proactive perception. However, integrating exteroception remains a significant challenge: existing approaches often fail to achieve high-speed, stable locomotion across diverse, unstructured terrains. In this work, we present \textit{Hiking in the Wild}, an extensible, end-to-end perceptive legged parkour framework designed for robust humanoid hiking. Our approach maps depth inputs and proprioception directly to joint actions, without relying on external localization and mapping. To ensure safety and training stability, we introduce two key mechanisms: a foothold safety mechanism combining extensible \textit{Terrain Edge Detection} with \textit{Foot Volume Points} to reduce catastrophic slippage on edges, and a \textit{Flat Patch Sampling} strategy that mitigates reward hacking by generating feasible navigation targets. Extensive experiments on the Unitree G1 humanoid robot demonstrate that our policy enables robust traversal of complex terrains at speeds up to 2.5 m/s. Code and video demonstrations are available on the project page: \url{https://project-instinct.github.io/hiking-in-the-wild}.
\end{abstract}


\section{Introduction}

Humanoid robots hold immense promise for traversing complex real-world environments and executing difficult tasks. Unlike their wheeled counterparts, humanoids can step over obstacles and cross discontinuous terrain. Recently, significant strides have been made in humanoid control, enabling robots to perform dynamic maneuvers such as dancing, backflips, and mimicking human motions \cite{liao2025beyondmimic, luo2025sonic}. However, there is a fundamental distinction between tracking a predefined motion and hiking in the wild. While tracking resembles memorizing a routine, hiking requires the robot to actively perceive the terrain, adapt to irregularities, and handle the unknown. In this paper, we study \textit{legged parkour}: agile traversal over obstacles using only legged contacts, without relying on arms, torso contacts, or full-body maneuvers such as vaulting and climbing.

Blind locomotion has established a robust baseline \cite{gu2024humanoid, radosavovic2024real, radosavovic2024learning, siekmann2021blind}, but its purely reactive nature is fundamentally limited: without perceiving ahead, the robot cannot safely handle deep gaps or high steps. Integrating exteroceptive perception remains challenging: LiDAR-based methods \cite{long2025learning, he2025attention, ben2025gallant} depend on precise state estimation and suffer from motion distortion at high speed, while depth-based methods that reconstruct heightmaps \cite{duan2024learning, sun2025dpl, song2025gait} are limited to low-speed, structured environments and struggle to scale to unseen wild terrains.

To address these challenges, we present \textit{Hiking in the Wild}, an extensible, end-to-end perceptive legged parkour framework designed for robust humanoid locomotion in unstructured environments. Unlike complex modular pipelines, our approach directly maps depth input and proprioception to joint actions. To handle the high dimensionality of visual data and the complexity of diverse terrain skills, we incorporate a Mixture-of-Experts (MoE) architecture \cite{jacobs1991adaptive, huang2025moe}. Our design allows the policy to capture depth inputs at high frequencies (up to 60 Hz). This high-bandwidth perception loop enables the robot to perform dynamic legged obstacle traversal and high-speed running (up to 2.5 m/s). Crucially, our system incorporates a realistic depth synthesis module that models sensor noise and artifacts during training, enabling zero-shot sim-to-real transfer without external mapping and localization of the environment.

A critical yet overlooked issue in humanoid RL is foothold precision: stepping partially on an edge can cause catastrophic slippage. Prior approaches \cite{zhuang2024humanoid, cheng2024extreme} require manually designed constraints for specific terrain types, limiting extensibility. We instead propose a geometric \textbf{Terrain Edge Detector} coupled with foot-attached \textbf{Volume Points} that automatically detects edges on arbitrary meshes and penalizes foot penetration, implicitly teaching the policy to seek safe, flat surfaces. Training stability is further challenged by ``reward hacking'', where random velocity commands lead the robot to spin rather than traverse obstacles \cite{cheng2024extreme}. We address this with a \textbf{Flat Patch Sampling} strategy that identifies reachable flat regions and derives speed-controlled commands toward them, ensuring the robot is always challenged with physically solvable tasks.

We validate our framework on the Unitree G1 humanoid robot, demonstrating robust traversal of stairs, ramps, gaps, and platforms at speeds up to 2.5~m/s. The contributions of this work are:

\begin{enumerate}
    \item An extensible, end-to-end perceptive \textbf{framework} for training and deployment based on depth input, enabling zero-shot sim-to-real transfer without external localization and mapping.
    \item A novel \textbf{safety mechanism} using Terrain Edge Detection and Foot Volume Points, alleviating the precise foothold problem in learning-based control.
    \item A \textbf{geometry-aware command generation} method that derives velocity commands by pointing toward reachable flat terrain patches, eliminating reward hacking and ensuring the robot actively traverses complex environments.
    \item Demonstration of robust wild hiking and legged parkour capabilities on a humanoid robot. The training and deployment code is \textbf{open-source and available to the community}, enabling deployment on real robots with minimal hardware modifications.
\end{enumerate}

\section{Related Work}

\subsection{Learning-based Legged Locomotion}
Legged locomotion control has shifted from model-based approaches, such as Model Predictive Control (MPC) \cite{wholebodyhumanoidmpcweb, scianca2020mpc, katayama2023model}, to data-driven reinforcement learning (RL) to better handle unmodeled dynamics. Early blind RL policies achieved impressive robustness on irregular terrains using only proprioception \cite{radosavovic2024real, siekmann2021blind, gu2024humanoid, yuan2025pvp}. However, these methods lack the exteroception required to navigate large obstacles quickly and safely.

To enable greater agility, recent works have integrated exteroceptive perception into the control loop \cite{cheng2024extreme, rudin2025parkour, wu2026perceptive, zhang2026rpl}. One category of methods relies on LiDAR to build elevation maps (2.5D) \cite{he2025attention, long2025learning, wang2025beamdojo, sun2025learning} or voxel grids \cite{ben2025gallant}. Nevertheless, these approaches depend heavily on precise localization, which limits their update frequency and overall robustness. Furthermore, LiDAR often suffers from motion distortion during fast movement, restricting these methods to low-speed scenarios. Alternatively, depth-based methods have been proposed. Some of these operate at low frequency \cite{zhuang2024humanoid}, while others employ intermediate modules to predict heightmaps \cite{duan2024learning, sun2025dpl, song2025gait} or learn end-to-end depth policies \cite{sun2026now, hao2026cref}, potentially compromising performance in high-speed or unseen wild environments. In contrast, our method directly utilizes high-frequency depth images as input to train an end-to-end policy, achieving highly dynamic behaviors across complex, unstructured terrains in the wild.

\subsection{Perceptive Foothold Control}
Explicit foothold control typically decouples perception and planning, utilizing terrain representations like elevation maps to solve for optimal placements. Approaches range from heuristic search \cite{wermelinger2016navigation, fankhauser2018robust, jenelten2020perceptive, kim2020vision, chilian2009stereo, mastalli2015line, agrawal2022vision} and learned feasibility costs \cite{magana2019fast, yang2021real, mastalli2017trajectory, wellhausen2021rough} to rigorous nonlinear optimization \cite{grandia2023perceptive, mastalli2020motion}. While hybrid frameworks that utilize probabilistic map uncertainty \cite{jenelten2020perceptive, fankhauser2018probabilistic} or combine pre-planned references with online tracking \cite{jenelten2024dtc, gangapurwala2021real, kim2025high} have achieved impressive agility, they remain brittle to state-estimation drift and reconstruction artifacts. Recently, end-to-end policies \cite{cheng2024extreme} use edge penalization to learn safe footholds implicitly. However, it is not extensible to arbitrary meshes. Our method takes a triangular mesh as input and automatically detects the edges, retaining the robustness of the implicit paradigm while improving safety.

\section{Method}

\begin{figure}[t]
  \centering
  \includegraphics[width=0.9\textwidth]{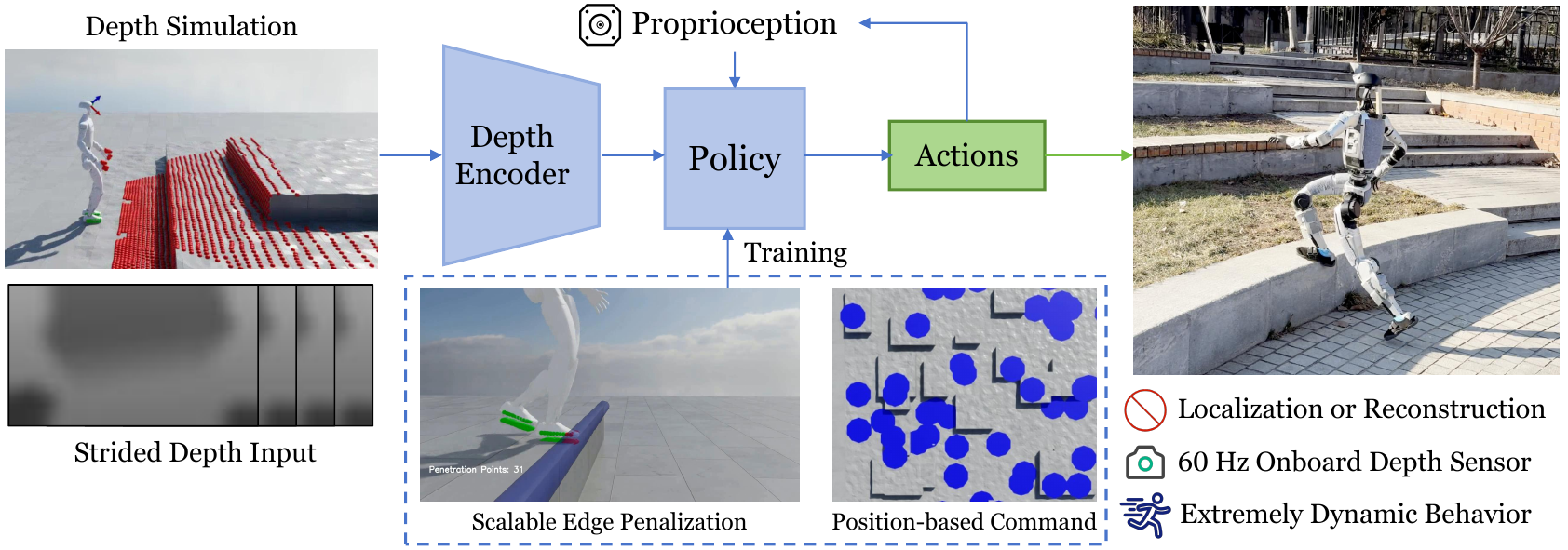}
  \vspace{-1mm}
  \caption{System overview. Our framework trains an end-to-end policy using simulated depth and proprioception. To ensure safety and agility on complex terrains, we incorporate Extensible Edge Penalization to avoid risky footholds and Position-Based Command Generation for precise tracking. The trained policy is directly deployed to the real robot (Zero-shot) using only a 60~Hz onboard depth camera as exteroception, achieving highly dynamic locomotion without explicit localization or map reconstruction.}
  \vspace{-3mm}
  \label{fig:overview}
\end{figure}

\subsection{Problem Formulation}

We formulate perceptive humanoid locomotion as a POMDP and train the policy with PPO \cite{schulman2017proximal}. The actor receives a history of $h$ proprioceptive states (angular velocity, projected gravity, velocity commands, joint positions/velocities, last action) concatenated with a strided depth history $\mathcal{H}_t$, with stochastic noise applied during training. During hardware deployment, velocity commands are provided by a human operator through a remote controller. An asymmetric critic additionally observes noise-free states and base linear velocity. The policy outputs target joint positions executed via PD control ($\boldsymbol{\tau}_t = k_p(\mathbf{a}_t - \mathbf{q}_t) - k_d\dot{\mathbf{q}}_t$) with gains from \cite{liao2025beyondmimic}. Episodes terminate on time-out, boundary violation, torso contact, or large base tilt (details in \autoref{app:termination}). The reward comprises task tracking, regularization, safety, and AMP-style terms; full specifications are in \autoref{tab:rewards_detail}.

\subsection{Ego-centric Depth Perception}
\label{fig:depth_sim}

\begin{figure}[t]
  \centering
  \includegraphics[width=\textwidth]{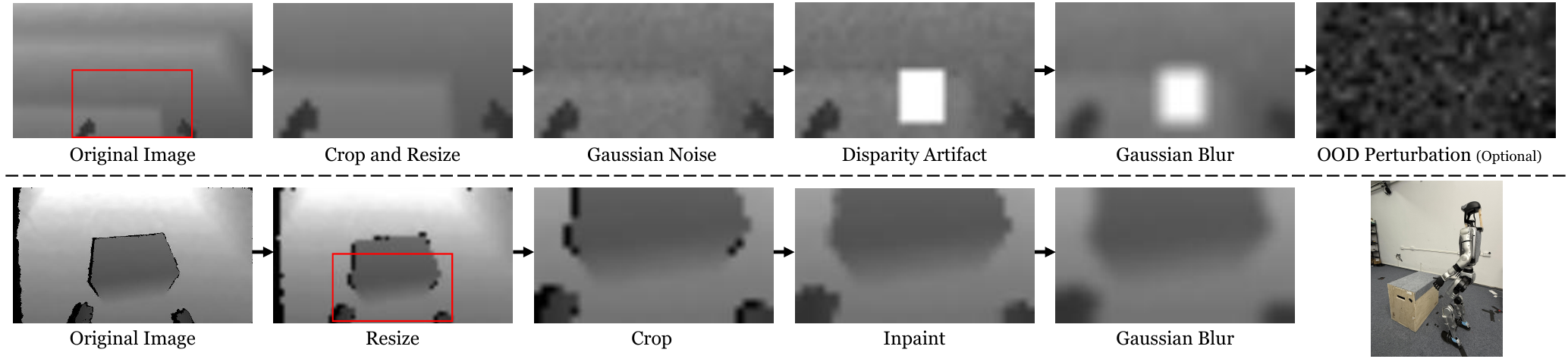}
  \vspace{-5mm}
  \caption{Depth processing: Top row shows $\mathcal{F}_{sim}$ on synthetic data, and bottom row shows $\mathcal{F}_{real}$ on real-world data.}
  \label{fig:depth_noise}
  \vspace{-6mm}
\end{figure}

We implement a GPU-accelerated ray-caster using NVIDIA Warp \cite{warp2022} to synthesize dense depth maps in real time. To bridge the sim-to-real gap, we define two alignment pipelines: $\mathcal{F}_{sim}$ degrades simulated depth via sequential stochastic operations (range-dependent Gaussian noise, disparity artifact synthesis, blur, and OOD perturbation), while $\mathcal{F}_{real}$ applies inpainting and blur to the physical sensor stream, harmonizing both distributions into a shared perception space (see \autoref{fig:depth_noise} and \autoref{app:depth_details}). The policy receives a depth history of $m$ frames with temporal stride $\ell$. During deployment, depth acquisition and preprocessing run asynchronously at 60~Hz, while the policy runs at 50~Hz using the latest available depth observations. History length and stride determine the look-back window without reducing the depth update rate. During training, a one-frame delay emulates sensor latency.

\subsection{Terrain Edge Contact Penalization}
\label{subsec:edge}
Driven by energy efficiency, RL agents naturally tend to minimize effort by taking shorter steps and often placing their feet on the nearest terrain edges \cite{cheng2024extreme}. While this strategy can yield high success rates in simulation due to simplified contact physics, it significantly increases the sim-to-real gap. In the real world, contact dynamics at structural edges are far more unpredictable than in simulation, and stepping partially on them frequently leads to catastrophic slippage and deployment failures. To mitigate this, we propose a soft-constraint mechanism that penalizes foot contacts near sharp geometric boundaries. First, we use a detector to find the sharp edges of the terrain mesh. Second, we attach a set of ``volume points'' to the robot's feet. By penalizing these points when they penetrate edge regions, the policy learns to select more stable, centered foot positions rather than risky ones.

\subsubsection{Terrain Edge Detector}

First, we identify sharp terrain edges by comparing the dihedral angle between adjacent faces to a predefined threshold $\tau$. To enhance computational efficiency, the resulting raw edges are further processed to filter noise and concatenate short segments. The edge processing step employs a greedy concatenation strategy to merge fragmented segments and reduce the total number of primitives. More details can be found in \autoref{appendix:edge_detector}.

\begin{figure}[t]
  \centering
  \includegraphics[width=\linewidth]{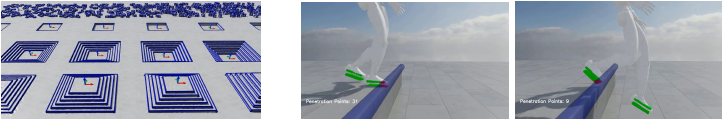}
  \caption{Terrain edge contact penalization. Left: automatically detected edges across diverse terrains. Right: volume points distributed within the foot manifold for penetration penalization.}
  \label{fig:edge_volume}
  \vspace{-3mm}
\end{figure}

\subsubsection{Volumetric Point Penetration Penalization}

To accurately monitor the contact state relative to terrain edges, we distribute a set of volume points $\mathcal{P}$ within each foot's collision geometry, as illustrated in \autoref{fig:edge_volume}. We leverage NVIDIA Warp \cite{warp2022} to perform massively parallel distance queries, computing the penetration depth of each point relative to the previously constructed spatial collision grid $\mathcal{S}$. 

To discourage unstable foot-ground interactions near terrain edges, we define a penalty term that considers both the geometric penetration and the foot velocity, following \cite{zhuang2023robot}. For each point $i \in \mathcal{P}$, let $\mathbf{d}_i$ be its penetration offset and $\mathbf{v}_i$ be its linear velocity in the world frame. The penalty term $r_{vol}$ is formulated as:
\begin{equation}
r_{vol} = \sum_{i=1}^{|\mathcal{P}|} \|\mathbf{d}_i\| \cdot (\|\mathbf{v}_i\| + \epsilon),
\end{equation}
where $\epsilon = 10^{-3}$ is a small constant for numerical stability. This formulation ensures that the policy is penalized more heavily for high-velocity impacts or scraping motions near terrain edges, thereby encouraging the robot to seek stable footholds.

\subsection{Position-based Velocity Command Generation}

During training, uniformly sampled velocity commands may cause ``reward hacking'' where the robot spins in place rather than traversing obstacles \cite{cheng2024extreme}. Prior fixes via reward shaping \cite{cheng2024quadruped} or goal-based commands \cite{zhang2024learning, zhu2025robust} either lack speed control or produce infeasible targets. We instead generate targets from reachable flat terrain patches and derive speed-controlled commands toward them.

We identify reachable flat regions on the terrain mesh via ray-casting: a candidate location is accepted as a navigation target only if the height variation within radius $r$ stays below threshold $\delta$, ensuring targets lie on stable ground rather than slopes or unreachable areas (see \autoref{alg:flat_patch} and \autoref{fig:flat_patches}). After generating the flat patches, we periodically select one as the navigation goal. The robot's velocity commands are then generated based on the relative position of this goal. Let $\mathbf{p}_{goal}^B = [x_g, y_g]^T$ represent the goal position in the robot's base frame. The desired linear velocity $v_x$ and angular velocity $\omega_z$ are calculated as follows:
\begin{align}
    v_x &= \text{clip}(k_v \cdot x_g, 0, v_{max}), \\
    \omega_z &= \text{clip}(k_{\omega} \cdot \text{atan2}(y_g, x_g), -\omega_{max}, \omega_{max}),
\end{align}
where $k_v$ and $k_{\omega}$ are the linear and angular stiffness coefficients. Because we use the forward camera on the robot's head link, we only focus on forward locomotion and heading alignment; therefore, the lateral velocity command $v_y$ is set to zero. The velocity limits $(v_{max}, \omega_{max})$ are adaptively adjusted based on the terrain category.

However, relying solely on these position-based commands can limit the robot's ability to learn in-place turning, as the goal is usually far away. To address this, we assign pure turning commands to a small subset of agents on flat terrain. For these agents, we set $v_x = 0$ and provide a random $\omega_z$, forcing the policy to learn effective in-place rotation maneuvers. This combination improves the overall maneuverability of the robot across different environments.

\subsection{Policy Training with Adversarial Motion Priors}

We adopt the AMP framework \cite{peng2021amp} to shape natural locomotion style. Walking and running policies use distinct reference datasets drawn from MPC-generated motion, retargeted human mocap, and running clips from LAFAN \cite{harvey2020robust}. We train the gaits separately because joint training produces fast walking with long strides rather than natural running. The discriminator operates on short state sequences rather than single transitions to better capture motion dynamics. Full dataset composition, discriminator formulation, and style-reward details are provided in \autoref{app:amp_details}.

\begin{figure}[t!]
  \centering
  \includegraphics[width=1.0\linewidth]{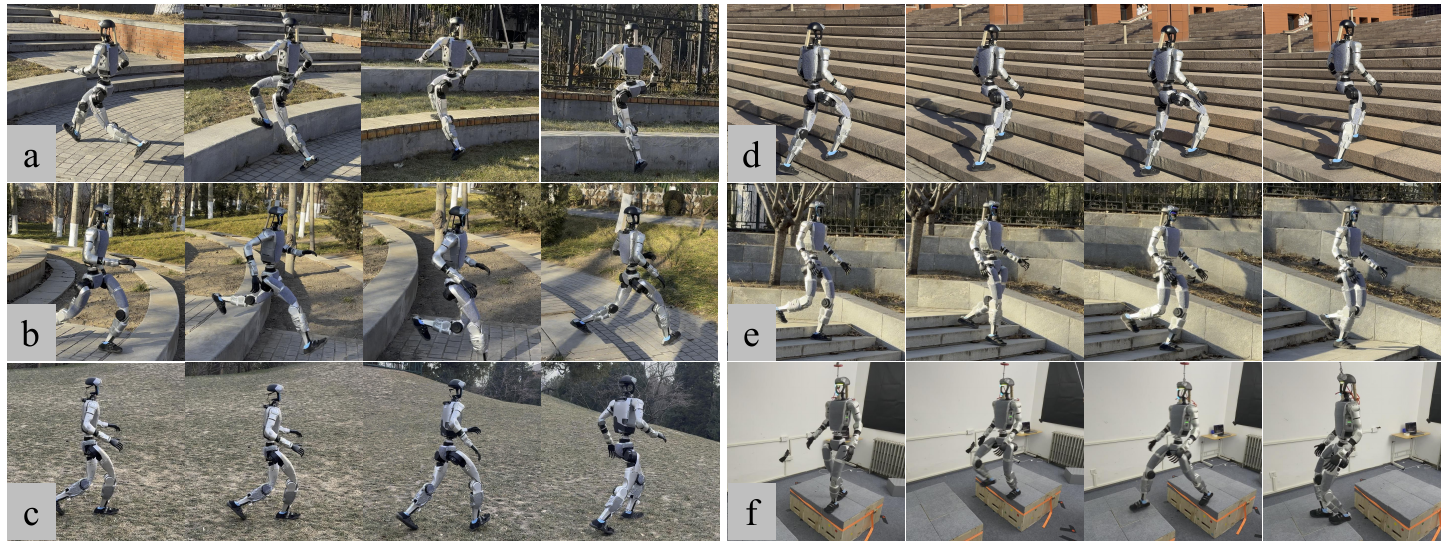}
  \vspace{-3mm}
  \caption{Snapshots of the humanoid robot traversing diverse indoor and outdoor environments via zero-shot transfer: (a) running ascent onto a high platform; (b) running descent from a high platform; (c) running on a grassy slope; (d) stair ascent; (e) stair descent; (f) traversing a discrete gap.}
  \label{fig:real-exp}
  \vspace{-5mm}
\end{figure}

\section{Experiments}

For the simulation results in \autoref{tab:edge_ablation} and \autoref{tab:ablation_result}, each method uses a single generalist walking policy across all evaluated terrains. Our HPL reimplementation is evaluated in the walking regime, and our running policy is excluded from this comparison because it does not cover all evaluated terrain types, including stairs. Hardware experiments use the deployment policies described in \autoref{sec:real_world_deployment} and \autoref{app:post_training}. Our evaluation addresses three questions:

\begin{itemize}
    \item \textbf{Q1}: Does our framework enable efficient training and reliable zero-shot deployment on physical hardware?
    \item \textbf{Q2}: Does the edge-aware penalization mechanism improve foothold safety and stability when traversing discrete terrain features? Is it extensible to new terrains without any extra design?
    \item \textbf{Q3}: What are the individual contributions of the key design components (e.g., perception, AMP, and command generation) to the robot's performance?
\end{itemize}

\subsection{Real-world Deployment}
\label{sec:real_world_deployment}

We deploy the trained policies onto the humanoid robot via zero-shot sim-to-real transfer. For hardware deployment, we use specialized policies for stair walking, platform walking, and running, while retaining the general walking policy for the remaining walking tasks (\autoref{app:post_training}). To switch policies, the robot first stops, transitions to a standing policy, and then switches to the selected locomotion policy. As illustrated in \autoref{fig:real-exp}, the robot successfully performs various locomotion primitives across challenging terrains. Notably, the robot achieves a maximum running speed of $2.5~\text{m/s}$ and successfully traverses difficult obstacles, including high platforms up to $32~\text{cm}$ and discrete gaps with a width of $50~\text{cm}$.  During these deployments, the policy uses depth observations acquired at 60~Hz to react to terrain changes.

To quantify the system's reliability, we conduct 10 trials for each evaluated terrain--gait combination. A trial is successful if the robot reaches the endpoint and stops stably; falling before completion is counted as a failure. \autoref{fig:real_success} reports the success rates for these combinations; absent bars indicate combinations that were not evaluated.

Furthermore, we conduct a long-duration test to evaluate the stability of the system over a prolonged period. The robot is tasked with continuous locomotion across multiple staircases and flat surfaces. It successfully maintains its balance and walks for 4~minutes without falls or physical assistance. This sustained performance highlights the effectiveness of our control framework and its potential for deployment in complex, long-range real-world tasks.

\subsection{Edge-aware Penalization Mechanism}
\label{sec:edge_evaluation}

\begin{figure}[b]
  \centering
  \vspace{-6mm}
  \includegraphics[width=0.8\linewidth]{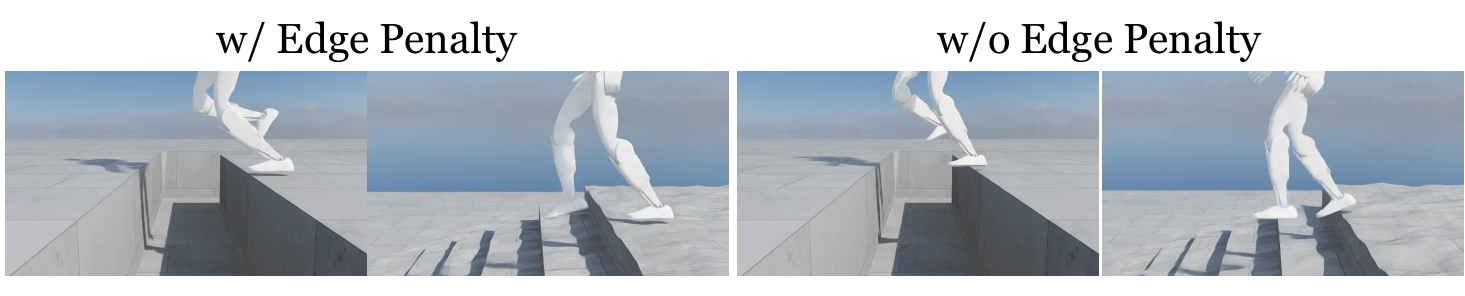}
  \vspace{-3mm}
  \caption{Visualization of foothold placement.}
  \label{fig:foothold}
\end{figure}

As introduced in \autoref{subsec:edge}, we utilize the \textit{Terrain Edge Detector} and \textit{Volumetric Point Penetration Penalization} to improve foothold safety. As illustrated in \autoref{fig:foothold}, with edge-aware penalization enabled, the robot tends to seek larger safety margins, effectively placing its feet away from terrain edges to maintain stable contact.

\begin{figure}[t]
  \centering
  \begin{minipage}[t]{0.42\linewidth}
    \vspace{0pt}
    \centering
    \includegraphics[width=\linewidth]{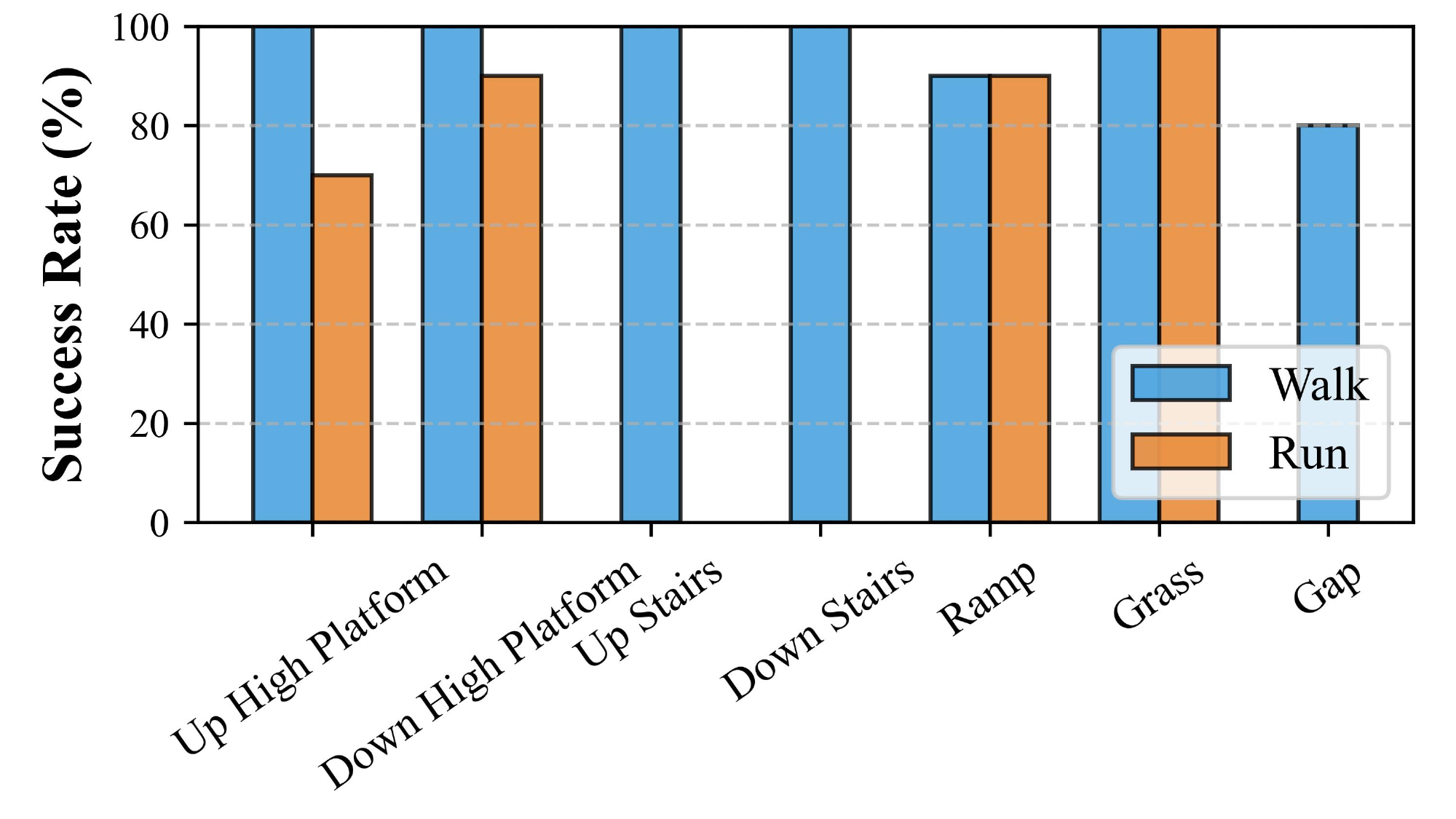}
  \end{minipage}\hfill
  \begin{minipage}[t]{0.54\linewidth}
    \vspace{2.5pt}
    \centering
    \scriptsize
    \setlength{\tabcolsep}{2.5pt}
    \renewcommand{\arraystretch}{1.45}
    \resizebox{\linewidth}{!}{%
    \begin{tabular}{lcccccc}
    \toprule
     & \multicolumn{2}{c}{\textbf{Sim SR $\uparrow$}} & \multicolumn{2}{c}{\textbf{Landing Area $\uparrow$}} & \multicolumn{2}{c}{\textbf{Real SR $\uparrow$}} \\
    \cmidrule(lr){2-3} \cmidrule(lr){4-5} \cmidrule(lr){6-7}
    \textbf{Terrain Type} & \textbf{Ours} & \textbf{No Edge} & \textbf{Ours} & \textbf{No Edge} & \textbf{Ours} & \textbf{No Edge} \\ \midrule
    Stair Up  & \underline{100.00\%} & \underline{100.00\%} & \textbf{0.77} & 0.68 & \textbf{10/10} & 8/10 \\
    Stair Down & \textbf{99.95\%} & 99.82\% & \textbf{0.67} & 0.50 & \textbf{10/10} & 6/10 \\
    Deep Gap     & \textbf{100.0\%}  & 99.94\% & \textbf{0.84} & 0.21 & \textbf{8/10} & 3/10 \\
    Small Box    & \textbf{99.09\%}  & 93.17\% & \textbf{0.81} & 0.70 & -- & -- \\
    \bottomrule
    \end{tabular}}
  \end{minipage}
  \par
  \begin{minipage}[t]{0.42\linewidth}
    \captionsetup{skip=0pt}
    \caption{Real-world success rates across terrains and gaits.}
    \label{fig:real_success}
  \end{minipage}\hfill
  \begin{minipage}[t]{0.54\linewidth}
    \captionsetup{skip=0pt}
    \captionof{table}{Edge penalization results in simulation and hardware.}
    \label{tab:edge_ablation}
  \end{minipage}
  \vspace{-2mm}
\end{figure}

To quantitatively evaluate this effect, we measure the \textbf{Success Rate} and \textbf{Mean Foot Landing Area Fraction} in simulation across various challenging terrains. The landing-area metric quantifies the fraction of the foot area supported by the terrain surface, reported on a scale from 0 to 1. To eliminate flat-ground dilution, it is computed exclusively when the robot's base is within the interaction range of the obstacle. We test each policy for 10,000 timesteps with 1,000 robots.
\autoref{tab:edge_ablation} compares our full method (\textit{Ours}) against the variant without edge penalization (\textit{No Edge}). Our method matches or exceeds \textit{No Edge} in simulation success rate and achieves a higher landing-area fraction across all tested terrains.

\begin{figure}[t!]
  \centering
  \includegraphics[width=\linewidth]{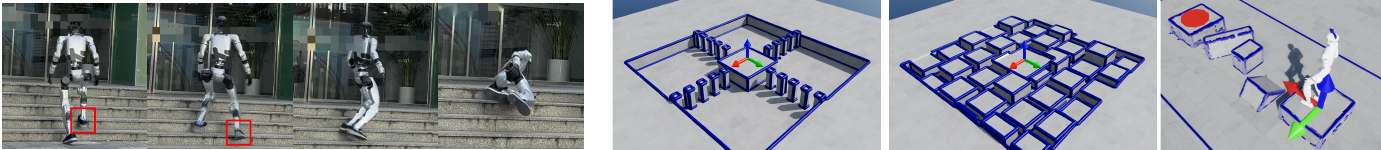}
  \caption{Left: the robot slips and falls after stepping on a stair edge. Right: edge detector visualizations on new and noisy terrains.}
  \label{fig:fall_more}
  \vspace{-4mm}
\end{figure}

In the real world, stepping on edges sometimes causes slippage or unpredictable contact dynamics, driving the robot into out-of-distribution (OOD) states that lead to immediate falls, as shown in \autoref{fig:fall_more}. By incentivizing the policy to maximize the landing area and avoid edges, our method significantly reduces these real-world risks, ensuring greater robustness during physical deployment. As shown in \autoref{fig:real-exp}, our policy exhibits ample safety margins and maintains stable footholds throughout the traversal.

To further demonstrate the extensibility of our framework, we evaluate the edge detector on new terrain types: \textit{Stones}, \textit{Stakes}, and \textit{Noisy Scan} terrain. As illustrated in \autoref{fig:fall_more}, the detector precisely identifies terrain edges without any manual feature engineering or parameter tuning. This robust zero-shot performance confirms that our edge-aware mechanism can scale to arbitrary terrain topologies.

\begin{table}[t!]
\centering
\caption{Quantitative results in simulation}
\label{tab:ablation_result}
\tiny
\setlength{\tabcolsep}{2.8pt}
\setlength{\heavyrulewidth}{0.05em}
\setlength{\lightrulewidth}{0.05em}
\setlength{\cmidrulewidth}{0.03em}
\resizebox{\linewidth}{!}{
\begin{tabular}{l ccccccccc c ccccccccc}
\toprule
 & \multicolumn{9}{c}{\textbf{Success Rate (\%) $\uparrow$}} & & \multicolumn{9}{c}{\textbf{Mean Reaching Time (s) $\downarrow$}} \\
\cmidrule(lr){2-10} \cmidrule(lr){12-20}
\textbf{Method} & \textbf{Large Box} & \textbf{Ramp} & \textbf{Small Box} & \textbf{Rough} & \textbf{Stair Down} & \textbf{Platform down} & \textbf{Stair Up} & \textbf{Platform up} & \textbf{Gap} & & \textbf{Large Box} & \textbf{Ramp} & \textbf{Small Box} & \textbf{Rough} & \textbf{Stair Down} & \textbf{Platform down} & \textbf{Stair Up} & \textbf{Platform up} & \textbf{Gap} \\
\midrule
w/o depth history  & 100.0 & 100.0 & 1.66 & 100.0 & 99.88 & 100.0 & 100.0 & 100.0 & 100.0 & & 6.20 & 6.72 & 11.68 & 5.78 & 5.33 & 5.81 & 6.54 & 5.89 & 5.79 \\
w/o position-based  & 23.13 & 100.0 & 90.68 & 100.0 & 99.59 & 100.0 & 95.18 & 11.70 & 100.0 & & 9.41 & 5.74 & 11.32 & 5.72 & 6.77 & 7.07 & 6.79 & 10.57 & 5.70 \\
w/o AMP & 13.75 & 99.94 & 0.00 & 100.0 & 98.39 & 99.47 & 0.00 & 0.00 & 99.94 & & 9.81 & \textbf{5.68} & -- & 6.50 & 6.89 & 7.40 & -- & -- & 6.07 \\
10Hz height map & 99.40 & 100.0 & 15.28 & 100.0 & 99.60 & 100.0 & 100.0 & 95.21 & 100.0 & & 6.51 & 6.01 & 14.73 & 6.34 & \textbf{4.40} & \textbf{5.63} & 5.63 & 6.03 & 6.40 \\
w/o depth & 38.94 & 100.0 & 3.33 & 100.0 & 99.12 & 99.93 & 94.54 & 0.00 & 67.95 & & 7.88 & 8.48 & 10.59 & 7.75 & 5.39 & 6.72 & 7.12 & -- & 7.58 \\
HPL~\cite{zhuang2024humanoid} & 43.65 & 99.71 & 51.01 & 100.0 & 99.90 & 100.0 & 92.07 & 97.84 & 99.95 & & 8.30 & 6.16 & 9.65 & 8.22 & 6.10 & 6.13 & 6.34 & 5.99 & 7.44 \\
\rowcolor{gray!20} Ours     & \textbf{100.0} & \textbf{100.0} & \textbf{99.09} & \textbf{100.0} & \textbf{99.95} & \textbf{100.0} & \textbf{100.0} & \textbf{100.0} & \textbf{100.0} & & \textbf{5.91} & 5.84 & \textbf{8.03} & \textbf{5.40} & \underline{5.15} & \textbf{5.63} & \textbf{5.36} & \textbf{5.60} & \textbf{5.43} \\
\bottomrule
\addlinespace
\end{tabular}
}
\vspace{-4mm}
\end{table}

\subsection{Comparison and Ablation}
\label{sec:comparison_ablation}

To evaluate both overall performance and the individual contributions of our proposed components, we compare our full framework against a reimplemented external baseline and several ablated variants:

\begin{itemize}
    \item \textbf{Single-frame Depth (w/o depth history)}: The policy receives only the current depth image instead of the strided temporal history.
    \item \textbf{Uniform Command (w/o position-based)}: The policy is trained using standard uniform velocity sampling with heading commands.
    \item \textbf{No Motion Prior (w/o AMP)}: The Adversarial Motion Prior is removed, training the policy solely through task-related rewards.
    \item \textbf{Blind Policy (w/o depth)}: The policy relies exclusively on proprioceptive feedback.
    \item \textbf{Height Map (10 Hz)}: We use a practical elevation-map update rate of 10~Hz~\cite{wang2025beamdojo}, compared with 60~Hz raw depth acquisition in our system.
    \item \textbf{HPL~\cite{zhuang2024humanoid}}: We reimplement the \textit{humanoid parkour learning} baseline in our training environment for comparison, as the original code is not open-sourced and the original work targets the Unitree H1 robot trained in IsaacGym.
\end{itemize}

We evaluate each policy with 1,000 robots in simulation and record their success rates and the mean reaching time. As shown in \autoref{tab:ablation_result}, our method achieves the highest success rates among both ablated variants and HPL, confirming the effectiveness of the complete framework on complex terrains. Our depth-based configuration outperforms the 10~Hz height-map baseline at their respective practical operating rates. AMP is an existing training component adopted in our framework. Removing it (\textit{w/o AMP}) reduces success rates to 0\% on Small Box, Stair Up, and Platform Up, indicating its importance for terrains requiring large leg lifts in our setting. The \textit{w/o position-based} variant degrades severely on Large Box (23.1\%) and Platform Up (11.7\%), confirming that position-based command generation is essential for directing the robot toward challenging obstacles. The blind policy (\textit{w/o depth}) fails on small and elevated structures where foothold precision requires visual feedback.

\section{Conclusion and Limitations}

In this paper, we present an extensible end-to-end perceptive locomotion framework for humanoid robots. By integrating a novel volumetric edge-aware penalization mechanism with position-based velocity command generation, our approach achieves highly dynamic behaviors on complex terrains. Extensive experiments in simulation and real-world deployment demonstrate that our framework enables humanoid robots to perform agile running and stable walking across a variety of challenging environments.

\textbf{Limitations.} A single forward-facing depth camera limits backward and lateral awareness. Hardware deployment uses specialized policies for stair walking, platform walking, and running; learning a unified policy remains challenging because joint training can cause mode collapse. Evaluation is limited to the Unitree G1 and static environments, with only 10 trials per evaluated terrain--gait combination. These results do not establish generalization across robot platforms or performance in dynamic environments. Failure-recovery capability is also limited. Future work includes broader perceptual coverage, unified policies, recovery behaviors, and more extensive hardware evaluation.

\clearpage
\acknowledgments{We thank Xiangting Meng, Baijun Ye, Siqiao Huang, and Yizhuo Gao for their assistance. This project is supported by Tsinghua University Initiative Scientific Research Program.}

\bibliography{references}

\clearpage
\appendix
\raggedbottom
\section*{Appendix}
\renewcommand{\thesection}{\Alph{section}}
\renewcommand{\thesubsection}{\Alph{section}.\arabic{subsection}}
\renewcommand{\thesubsubsection}{\Alph{section}.\arabic{subsection}.\arabic{subsubsection}}

\noindent\textbf{Table of Contents}

\noindent\textbf{A. Policy Training Details}\\
\hspace*{1.5em} A.1\ \ Policy Observation, Action, and Noise\\
\hspace*{1.5em} A.2\ \ Termination Criteria\\
\hspace*{1.5em} A.3\ \ Reward Function\\
\hspace*{1.5em} A.4\ \ Ego-centric Depth Perception Details\\
\hspace*{1.5em} A.5\ \ Flat Patch Sampling\\
\hspace*{1.5em} A.6\ \ Terrain Configuration\\
\hspace*{1.5em} A.7\ \ Policy Network\\
\hspace*{1.5em} A.8\ \ Post-Training for Hardware Deployment\\
\hspace*{1.5em} A.9\ \ AMP Training Details\\
\hspace*{1.5em} A.10 Terrain Edge Detector Algorithm\\
\noindent\textbf{B. Experiment Configurations}\\
\noindent\textbf{C. Simulation Evaluation Details}

\vspace{4mm}

\section{Policy Training Details}
\label{app:policy_training_details}

\subsection{Policy Observation, Action, and Noise}

The actor and critic observations are summarized in \autoref{tab:mdp_spec}. The policy uses an asymmetric actor-critic architecture: the actor receives noisy proprioceptive states and depth history; the critic additionally observes noise-free states and base linear velocity. All proprioceptive signals use a history of $h=8$ steps. Noise scales and distributions for each term are provided in \autoref{tab:obs_noise}.

\begin{table}[h]
\centering
\small
\caption{Observation and action space specification.}
\label{tab:mdp_spec}
\renewcommand{\arraystretch}{1.2}
\setlength{\tabcolsep}{4pt}
\begin{tabular}{l l c l}
\toprule
\textbf{Space} & \textbf{Signal} & \textbf{Dim} & \textbf{Description} \\
\midrule
\multirow{7}{*}{\textbf{Actor Obs}}
  & Base Angular Velocity $\boldsymbol{\omega}_t$ & $3$ & IMU gyroscope \\
  & Projected Gravity $\mathbf{g}_t$ & $3$ & Gravity direction in base frame \\
  & Velocity Commands $\mathbf{c}_t$ & $3$ & $(v_x, v_y, \omega_z)$ \\
  & Joint Positions $\mathbf{q}_t$ & $29$ & 29-DoF joint angles \\
  & Joint Velocities $\dot{\mathbf{q}}_t$ & $29$ & 29-DoF joint velocities \\
  & Last Action $\mathbf{a}_{t-1}$ & $29$ & Previous target joint positions \\
  & Depth History $\mathcal{H}_t$ & $m{\times}H{\times}W$ & Strided depth frames ($32{\times}18$ each) \\
\midrule
\textbf{Critic Obs} & + Base Linear Velocity $\mathbf{v}_t$ & $3$ & Noise-free; critic only \\
\midrule
\textbf{Action} & Target Joint Positions $\mathbf{a}_t$ & $29$ & Executed via PD control: $\boldsymbol{\tau}_t = k_p(\mathbf{a}_t - \mathbf{q}_t) - k_d\dot{\mathbf{q}}_t$ \\
\bottomrule
\end{tabular}
\end{table}

The detailed noise configuration for each proprioceptive term is summarized in \autoref{tab:obs_noise}. The detailed exteroceptive depth observation is illustrated in \autoref{fig:depth_sim}.

\begin{table}[h]
\centering
\small
\caption{Detailed policy proprioceptive observation space, scaling factors, and noise models.}
\label{tab:obs_noise}
\renewcommand{\arraystretch}{1.2}
\setlength{\tabcolsep}{4pt}
\begin{tabular}{l c c c}
\toprule
\textbf{Observation Term} & \textbf{Scale} & \textbf{Noise Type} & \textbf{Noise Parameters} \\ 
\midrule
Base Angular Velocity ($\omega_t$) & $0.25$ & Uniform & $\mathcal{U}(-0.2, 0.2)$ rad/s \\
Projected Gravity ($\mathbf{g}_t$) & $1.0$ & Uniform & $\mathcal{U}(-0.05, 0.05)$ \\
Velocity Commands ($\mathbf{c}_t$) & $1.0$ & None & N/A \\
Joint Positions ($q_t$) & $1.0$ & Uniform & $\mathcal{U}(-0.01, 0.01)$ rad \\
Joint Velocities ($\dot{q}_t$) & $0.05$ & Uniform & $\mathcal{U}(-0.5, 0.5)$ rad/s \\
Last Action ($a_{t-1}$) & $1.0$ & None & N/A \\
\bottomrule
\end{tabular}
\end{table}

\subsection{Termination Criteria}
\label{app:termination}

To ensure stable and efficient training, the training episode terminates immediately if any of the following conditions are met. These criteria prevent the policy from exploring physically invalid states or dangerous configurations. The specific thresholds and parameters used in our MDP are detailed below:

\begin{itemize}
    \item \textbf{Time Out:} The episode terminates naturally if the maximum episode duration is reached.
    \item \textbf{Terrain Out-of-Bounds:} The episode ends if the robot moves beyond the valid terrain boundaries. A distance buffer of $2.0$~m is applied to allow for minor deviations at the boundaries.
    \item \textbf{Illegal Contact:} To penalize falls, the episode terminates if the robot's torso (\texttt{torso\_link}) experiences a contact force exceeding a threshold of $1.0$~N.
    \item \textbf{Bad Orientation:} To maintain stability, the episode ends if the tilt angle of the robot's base exceeds $1.0$~rad relative to the gravity vector.
    \item \textbf{Insufficient Root Height:} The episode terminates if the robot's root height drops below $0.5$~m relative to the environment origin, indicating a fall or a failure to stand up.
\end{itemize}

\subsection{Reward Function}

\label{app:rewards}

The reward function is designed to encourage velocity tracking while enforcing physical safety. It consists of four components: task reward, regularization reward, safety reward, and AMP-style reward. The detailed definition of each term, along with its corresponding weight and key parameters, is provided in \autoref{tab:rewards_detail}.

\begin{table}[ht]
\centering
\small
\caption{Detailed specifications of reward terms, weights, and mathematical formulations.}
\label{tab:rewards_detail}
\renewcommand{\arraystretch}{1.5}
\setlength{\tabcolsep}{6pt}
\resizebox{\linewidth}{!}{%
\begin{tabular}{l c l}
\toprule
\textbf{Reward Term} & \textbf{Weight} & \textbf{Mathematical Formulation} \\
\midrule
\multicolumn{3}{l}{\textit{\textbf{Task Reward}}} \\
Linear Vel Tracking ($v_{xy}$) & $2.0$ & $\exp(-\| \mathbf{v}_{xy}^* - \mathbf{v}_{xy} \|^2 / 0.5^2)$ \\
Angular Vel Tracking ($\omega_z$) & $2.0$ & $\exp(-(\omega_z^* - \omega_z)^2 / 0.5^2)$ \\
Heading Error & $-1.0$ & $|\omega_z^*|$ \\
Don't Wait & $-0.5$ & $\mathbb{I}(v_x^* > 0.3) \cdot (\mathbb{I}(v_x < 0.15) + \mathbb{I}(v_x < 0) + \mathbb{I}(v_x < -0.15))$ \\
Is Alive & $3.0$ & $+1$ \\
Stand Still & $-0.3$ & $(\|\mathbf{q} - \mathbf{q}_{default}\|_1 - 4.0) \cdot \mathbb{I}(\|\mathbf{v}^*\| < 0.15) \cdot \mathbb{I}(|\omega_z^*| < 0.15)$ \\
\midrule
\multicolumn{3}{l}{\textit{\textbf{Regularization Reward}}} \\
Edge Penetration & $-4.0$ & $\sum_{i=1}^{|\mathcal{P}|} \|\mathbf{d}_i\| \cdot (\|\mathbf{v}_i\| + \epsilon)$ \\
Feet Air Time & $0.5$ & $\min_{f} (t_{phase, f}) \cdot \mathbb{I}(\sum c_f = 1) \cdot \mathbb{I}(\|\mathbf{v}^*\| > 0.15)$ \\
Feet Slide & $-0.4$ & $\sum_{f} \|\mathbf{v}_{xy, f}\| \cdot \mathbb{I}(c_f)$ \\
Joint Deviation (Hip) & $-0.5$ & $\sum_{j \in \text{hips}} (q_j - q_{j, default})^2$ \\
Base Ang Vel (XY) ($L_2$) & $-0.05$ & $\| \omega_{xy} \|^2$ \\
Joint Torques ($L_2$) & $-1.5\text{e-}7$ & $\| \boldsymbol{\tau}_{legs} \|^2$ \\
Joint Acc ($L_2$) & $-1.25\text{e-}7$ & $\| \ddot{\mathbf{q}} \|^2$ \\
Joint Vel ($L_2$) & $-1.0\text{e-}4$ & $\| \dot{\mathbf{q}} \|^2$ \\
Action Rate ($L_2$) & $-0.005$ & $\| \mathbf{a}_t - \mathbf{a}_{t-1} \|^2$ \\
Flat Orientation & $-3.0$ & $\| \mathbf{g}_{xy}^{proj} \|^2$ \\
Pelvis Orientation & $-3.0$ & $\| \mathbf{g}_{xy}^{proj, pelvis} \|^2$ \\
Feet Orientation & $-0.4$ & $\sum_{f} \| \mathbf{g}_{xy, f}^{proj} \| \cdot \mathbb{I}(c_f)$ \\
Feet Height Error & $-0.1$ & $\sum_{f} \sum_{p} \text{clip}(h_{f} - h_{terr, p} - 0.035, 0, 0.3) \cdot \mathbb{I}(c_f)$ \\
Feet Distance & $0.4$ & $\exp(-\max(0, 0.12 - |y_{L}^R - y_{R}^R|) / 0.05) - 1$ \\
Energy Consumption & $-5.0\text{e-}5$ & $\sum_{j} (\tau_j \dot{q}_j / k_j)^2$ \\
Freeze Upper Body & $-0.004$ & $\|\mathbf{q}_{upper} - \mathbf{q}_{upper}^{default}\|_1$ \\
\midrule
\multicolumn{3}{l}{\textit{\textbf{Safety Reward}}} \\
Joint Pos Limits & $-1.0$ & $\sum_{j} (\max(0, q_j - q_{j, max}) + \max(0, q_{j, min} - q_j))$ \\
Joint Vel Limits & $-1.0$ & $\sum_{j} \max(0, |\dot{q}_j| - 0.9\dot{q}_{j, max})$ \\
Torque Limits & $-0.01$ & $\sum_{j} \max(0, |\tau_j| - 0.8\tau_{j, max})^2$ \\
Undesired Contacts & $-1.0$ & $\mathbb{I}(\text{count}(\text{collision}_{body \setminus feet}) > 0)$ \\
\midrule
\multicolumn{3}{l}{\textit{\textbf{AMP Reward}}} \\
AMP Style & $0.25$ & $\max\left[ 0, 1 - 0.25(D(\mathbf{S}_t) - 1)^2 \right]$ \\
\bottomrule
\end{tabular}%
}
\end{table}

\subsection{Ego-centric Depth Perception Details}
\label{app:depth_details}

\subsubsection{Efficient Depth Simulation via Parallelized Ray-casting}

To synthesize high-fidelity depth observations, we leverage the NVIDIA Warp framework \cite{warp2022} to implement a GPU-accelerated ray-caster. For each pixel $(i, j)$ in the image plane, a ray is emitted in direction $\mathbf{v}_{i,j}$. The ray-caster computes the radial distance $d_{i,j}$ by identifying the first intersection with the scene geometry $\mathcal{G}$:
\begin{equation}
    d_{i,j} = \min \{ \tau \mid \mathbf{p}_c + \tau \mathbf{v}_{i,j} \cap \mathcal{G} \neq \emptyset \}.
\end{equation}
The orthogonal depth $z_{i,j}$ is obtained by projecting onto the camera's principal axis $\mathbf{n}_c$:
\begin{equation}
    z_{i,j} = d_{i,j} \cdot (\mathbf{v}_{i,j} \cdot \mathbf{n}_c).
\end{equation}

\subsubsection{Bidirectional Depth Alignment and Noise Modeling}

We define two transformation pipelines that map depth from each domain into a unified perception space $\mathcal{O}$, targeting $P(\mathcal{F}_{sim}(d_{sim})) \approx P(\mathcal{F}_{real}(d_{real}))$.

\paragraph{Simulation Pipeline $\mathcal{F}_{sim}$}
Sequential stochastic operations applied to simulated depth:
\begin{enumerate}
    \item \textbf{Crop and Resize}: Rescaled to target resolution, focusing on the terrain-relevant bottom-center region.
    \item \textbf{Range-dependent Gaussian Noise}: Additive noise $\epsilon \sim \mathcal{N}(0, \sigma^2)$ injected within valid sensing range $[d_{min}, d_{max}]$:
    $z'_{i,j} = z_{i,j} + \epsilon \cdot \mathbb{I}[z_{i,j} \in [d_{min}, d_{max}]]$.
    \item \textbf{Disparity Artifact Synthesis}: Contiguous pixels masked as invalid to simulate binocular matching failures.
    \item \textbf{Gaussian Blur}: Convolved with blur kernel $\mathbf{K}$ to simulate motion blur.
    \item \textbf{Clipping and Normalization}: Values clipped and normalized to $[0, 1]$.
    \item \textbf{Out-of-distribution (OOD) Perturbation}: With probability $P_{ood}$, the entire observation is replaced by Gaussian noise to handle transient perceptual failures.
\end{enumerate}

\paragraph{Real-world Pipeline $\mathcal{F}_{real}$}
Applied to the physical depth stream during deployment:
\begin{enumerate}
    \item \textbf{Crop and Resize}: Aligned to simulation input geometry.
    \item \textbf{Depth Inpainting}: Recovers zero-valued pixels caused by occlusion or disparity shadows.
    \item \textbf{Gaussian Blur}: Suppresses high-frequency sensor jitter.
\end{enumerate}

\subsubsection{Temporal Depth Aggregation via Strided Sampling}

The policy receives a history buffer of $m$ depth frames sampled at temporal stride $\ell$ and delayed by one frame to emulate sensor latency:
\begin{equation}
    \mathcal{H}_t = \{ I_{t - k \cdot \ell - 1} \mid k = 0, 1, \dots, m-1 \}.
\end{equation}
This representation spans $(m-1)\cdot\ell$ steps using $m$ frames. The stride spaces the historical samples without reducing the update rate of the newest frame.

\subsection{Flat Patch Sampling}
\label{app:flat_patch}

\begin{algorithm}[H]
\caption{Flat Patch Sampling}
\label{alg:flat_patch}
\SetKwInOut{Input}{Input}
\SetKwInOut{Output}{Output}
\Input{Mesh $\mathcal{M}$, Patch radius $r$, Max height difference $\delta$}
\Output{Set of valid targets $\mathcal{P}$}
\BlankLine
\While{$|\mathcal{P}| < N_{targets}$}{
Sample a random 2D position $(x, y)$ in the environment\;
Ray-cast to get a set of heights $H$ within radius $r$ around $(x, y)$\;
\If{$\max(H) - \min(H) < \delta$}{Add $(x, y, \text{avg}(H))$ to $\mathcal{P}$\;}}
\Return{$\mathcal{P}$}\;
\end{algorithm}

\begin{figure}[H]
  \centering
  \includegraphics[width=\linewidth]{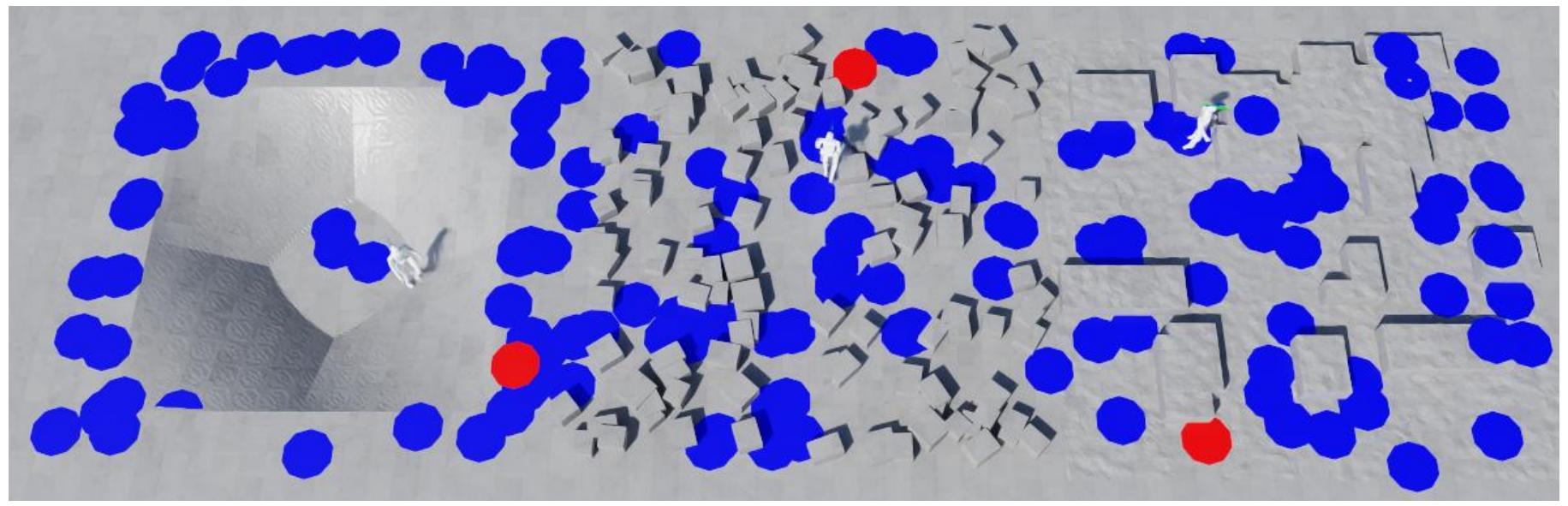}
  \caption{Flat patches on different terrains.}
  \label{fig:flat_patches}
\end{figure}

Unless otherwise specified, the command-generation and flat-patch parameters are $\epsilon=1\times10^{-6}$, $k_v=2.0$, $k_\omega=2.0$, target radius $r=0.2$, and flatness threshold $\delta=0.05$.

\subsection{Terrain Configuration}
\label{app:terrain_config}

To ensure the robot learns robust locomotion skills across diverse environments, we employ a curriculum-based terrain generation pipeline. The training environment consists of a grid of $10 \times 20$ sub-terrains, each measuring $8.0 \times 8.0$ meters. The height field resolution is set to $0.05$~m horizontally and $0.005$~m vertically.

The detailed specifications for each terrain type, including their geometric parameters, curriculum proportions, and the associated velocity command ranges used for training, are summarized in \autoref{tab:terrain_combined}. We randomly generate walls around the terrains to simulate an indoor environment with probability $\mathcal{P}=0.3$. Note that we use Rough (Stand) to train the robot for in-place turning.

\subsection{Policy Network}
The learning framework consists of three primary components: a CNN-based visual encoder, a Mixture-of-Experts (MoE) policy, and an Adversarial Motion Prior (AMP) discriminator. The visual encoder processes the strided depth history through a convolutional network, followed by a projection MLP to extract a visual embedding. This embedding is concatenated with the proprioceptive observations and fed into the MoE-based actor-critic network, where a gating mechanism dynamically weights the contributions of multiple expert networks to generate robust actions. Additionally, the AMP discriminator distinguishes between the actor's generated motions and the reference dataset. The detailed hyperparameters for these components are listed in \autoref{tab:net_arch_params}.

\begin{table}[H]
\centering
\small
\caption{Detailed hyperparameters for the network architecture.}
\label{tab:net_arch_params}
\renewcommand{\arraystretch}{1.2}
\setlength{\tabcolsep}{12pt}
\begin{tabular}{c|c}
\toprule
\textbf{Parameter} & \textbf{Value} \\
\midrule
CNN channels & $[4]$ \\
CNN kernel sizes & $[3]$ \\
CNN strides & $[1]$ \\
CNN paddings & $[1]$ \\
CNN pooling & MaxPool \\
Encoder projection units & $[256, 256]$ \\
Encoder embedding dim & $128$ \\
Encoder activation & ReLU \\
\midrule
Policy Type & MoE \\
Number of Experts & $4$ \\
Gating hidden units & $[256]$ \\
Expert hidden units & $[256, 128, 64]$ \\
Policy activation & ELU \\
\midrule
Discriminator hidden units & $[1024, 512]$ \\
Discriminator activation & ReLU \\
\bottomrule
\end{tabular}
\end{table}

\begin{table}[ht]
\centering
\caption{Detailed specifications of terrain types and their corresponding velocity command ranges for walking and running modes. Note that the lateral velocity command $v_y$ is set to $0$ for all terrain types.}
\label{tab:terrain_combined}
\renewcommand{\arraystretch}{1.2}
\setlength{\tabcolsep}{4pt}
\resizebox{\linewidth}{!}{%
\begin{tabular}{l l c c c c c c}
\toprule
\multicolumn{2}{c}{\textbf{Terrain Settings}} & \multicolumn{3}{c}{\textbf{Command Range (walking)}} & \multicolumn{3}{c}{\textbf{Command Range (run)}} \\
\cmidrule(lr){1-2}\cmidrule(lr){3-5}\cmidrule(lr){6-8}
\textbf{Terrain Type} & \textbf{Geometric Parameters} & \textbf{Prop.} & $\mathbf{v_x}$ \textbf{(m/s)} & $\mathbf{\omega_z}$ \textbf{(rad/s)} & \textbf{Prop.} & $\mathbf{v_x}$ \textbf{(m/s)} & $\mathbf{\omega_z}$ \textbf{(rad/s)} \\ 
\midrule
Rough & Perlin Noise Scale $\in [0.0, 0.1]$~m & $5\%$ & $[0.45, 1.0]$ & $[-1.0, 1.0]$ & $20\%$ & $[1.95, 3.05]$ & $[-1.0, 1.0]$ \\
Rough (Stand) & Perlin Noise Scale $\in [0.0, 0.1]$~m & $5\%$ & $0.0$ & $[-1.0, 1.0]$ & $5\%$ & $0.0$ & $[-1.0, 1.0]$ \\
Square Gaps & Gap Dist $\in [0.1, 0.7]$~m, Depth $\in [0.4, 0.6]$~m & $10\%$ & $[0.45, 0.8]$ & $[-1.0, 1.0]$ & $-$ & $-$ & $-$ \\
Stairs Down & Step Height $\in [0.05, 0.23]$~m, Width $0.3$~m & $15\%$ & $[0.45, 0.8]$ & $[-1.0, 1.0]$ & $-$ & $-$ & $-$ \\
Platform Down & Platform Height $\in [0.05, 0.45]$~m & $10\%$ & $[0.45, 0.8]$ & $[-1.0, 1.0]$ & $25\%$ & $[1.95, 3.05]$ & $[-1.0, 1.0]$ \\
Stairs Up & Step Height $\in [0.05, 0.23]$~m, Width $0.3$~m & $15\%$ & $[0.45, 0.8]$ & $[-1.0, 1.0]$ & $-$ & $-$ & $-$ \\
Platform Up & Platform Height $\in [0.05, 0.45]$~m & $10\%$ & $[0.45, 0.8]$ & $[-1.0, 1.0]$ & $25\%$ & $[1.95, 3.05]$ & $[-1.0, 1.0]$ \\
Discrete Boxes & Height $\in [0.05, 0.45]$~m, Width $\in [0.8, 1.5]$~m & $10\%$ & $[0.45, 0.8]$ & $[-1.0, 1.0]$ & $-$ & $-$ & $-$ \\
Mesh Boxes & Height Mean $\in [0.1, 0.4]$~m & $10\%$ & $[0.45, 0.8]$ & $[-1.0, 1.0]$ & $-$ & $-$ & $-$ \\
Slopes & Slope $\in [0.0, 0.7]$ rad & $10\%$ & $[0.45, 0.8]$ & $[-1.0, 1.0]$ & $25\%$ & $[1.95, 3.05]$ & $[-1.0, 1.0]$ \\
\bottomrule
\end{tabular}%
}
\end{table}

\subsection{Post-Training for Hardware Deployment}
\label{app:post_training}

The general walking policy is trained across multiple terrain categories and serves as the initialization for hardware post-training. To improve hardware stability, we fine-tune three specialized policies from this checkpoint: \textit{stair walking}, \textit{platform walking}, and \textit{running}. Each specialized policy is trained for an additional $40{,}000$ iterations. During hardware deployment, the specialized policies handle their respective tasks, while the general walking policy handles the remaining walking tasks.

Post-training is conducted on flat ground combined with the relevant target terrain. To focus each policy on its intended operating regime, we narrow the velocity command range compared to general training: the walking policies use $v_x \in [0.45, 0.55]$~m/s and the running policy uses $v_x \in [1.95, 3.05]$~m/s. The lateral command is fixed at $v_y = 0$ for all post-training runs. All other training hyperparameters remain identical to those used in the general policy.

\subsection{AMP Training Details}
\label{app:amp_details}

\subsubsection{Reference Dataset Composition}

The reference dataset $\mathcal{D}$ is collected at $50$~Hz from three sources:
\begin{enumerate}
    \item \textbf{Synthetic Data}: Walking motions from an MPC controller \cite{wholebodyhumanoidmpcweb}.
    \item \textbf{Human Motion}: Mocap data captured via the NOKOV system, including stair and platform traversal, retargeted to the robot via GMR \cite{araujo2025retargeting}.
    \item \textbf{Running Motion}: High-speed running clips from the LAFAN dataset \cite{harvey2020robust}.
\end{enumerate}
To avoid mode collapse, walking and running policies use separate datasets: $\mathcal{D}_{\text{walk}}$ combines sources 1 and 2 ($T = 379.62$~s); $\mathcal{D}_{\text{run}}$ uses source 3 ($T = 1.54$~s, one gait cycle). Symmetric augmentation is applied to all datasets. Joint training produces fast walking with long strides rather than natural running. We attribute this to the lack of transition data between walking demonstrations (up to 1~m/s) and running demonstrations (above 2~m/s), together with the smaller amount of running data. The running policy is therefore initialized from the general walking checkpoint and trained separately using $\mathcal{D}_{\text{run}}$, as described in \autoref{app:post_training}. Despite the narrow running distribution, the policy naturally generalizes to turning and climbing at inference time. This suggests that AMP requires as little as a single complete gait cycle to effectively imitate a target motion style, making it practical to deploy with minimal reference data.

\subsubsection{Discriminator and Style Reward}

The discriminator $D(\mathbf{S})$ receives short temporal sequences rather than single state pairs, enabling it to capture motion dynamics. It is trained with a least-squares (MSE) loss:
\begin{equation}
    L_D = \mathbb{E}_{\mathcal{M}}[(D(\mathbf{S}) - 1)^2] + \mathbb{E}_{\mathcal{P}}[(D(\mathbf{S}) + 1)^2],
\end{equation}
where $\mathcal{M}$ is the reference dataset and $\mathcal{P}$ is the policy rollout. Gradient penalty and weight decay are applied for training stability. The style reward fed back to the policy is:
\begin{equation}
    r_t = \max\left[ 0, 1 - 0.25(D(\mathbf{S}_t) - 1)^2 \right].
\end{equation}
Compared to binary cross-entropy and log-based rewards, this combination of an MSE loss and a quadratic reward yields smoother, non-saturating gradients, improving convergence toward the reference motion manifold.

\subsection{Terrain Edge Detector Algorithm}
\label{appendix:edge_detector}

We first identify sharp terrain edges by comparing the dihedral angle between adjacent faces to a predefined threshold $\tau$, as shown in \autoref{alg:edge_cylinder_gen}.

\begin{algorithm}[H]
\SetKwInOut{Input}{Input}
\SetKwInOut{Output}{Output}
\SetKwFunction{ProcessEdges}{ProcessEdges}
\SetKwFunction{BuildGrid}{CylinderSpatialGrid}
\SetKwFunction{DegToRad}{Deg2Rad}
\caption{Terrain Edge Detection Algorithm}
\label{alg:edge_cylinder_gen}
\Input{
    Triangular Mesh $\mathcal{M} = (\mathcal{V}, \mathcal{F})$, 
    Sharpness Threshold $\tau$, 
    Cylinder Radius $r$, 
    Grid Resolution $N_{\text{grid}}$
}
\Output{Spatial Collision Grid $\mathcal{S}$}
\BlankLine

\tcc{1. Sharp Edge Detection}
Let $\mathcal{A}$ be the set of face adjacencies in $\mathcal{M}$\;
Initialize raw edge set $E_{\text{raw}} \leftarrow \emptyset$\;
\ForEach{adjacency $a \in \mathcal{A}$}{
    Let $\alpha_a$ be the dihedral angle of $a$\;
    \If{$\alpha_a > \tau$}{
        Let $(v_i, v_j)$ be the vertex indices shared by $a$\;
        Add the segment coordinates $(\mathcal{V}[v_i], \mathcal{V}[v_j])$ to $E_{\text{raw}}$\;
    }
}
\BlankLine
\tcc{2. Edge Processing}
\eIf{$E_{\text{raw}} \text{ is empty}$}{
    \Return{None}\;
}{
    $E_{\text{final}} \leftarrow \ProcessEdges(E_{\text{raw}})$\;
}
\BlankLine

\tcc{3. Spatial Structure Construction}
Initialize cylinder set $\mathcal{C} \leftarrow \emptyset$\;
\ForEach{segment $e \in E_{\text{final}}$}{
    Construct cylinder $c$ from segment $e$ with radius $r$\;
    Add $c$ to $\mathcal{C}$\;
}

$\mathcal{S} \leftarrow \BuildGrid(\mathcal{C}, N_{\text{grid}})$\;

\Return{$\mathcal{S}$}\;
\end{algorithm}

Then, we employ a greedy edge concatenation process to refine raw edges, as shown in \autoref{alg:greedy_edge}. By merging adjacent collinear segments into single extended edges, this approach significantly reduces storage requirements and reduces computation time.

\begin{algorithm}[H]
\SetKwInOut{Input}{Input}
\SetKwInOut{Output}{Output}

\caption{Greedy Edge Concatenation Process}
\label{alg:greedy_edge}

\Input{Edge set $E_{raw}$, Angle threshold $\theta$, Min points $k_{\min}$, Tol $\epsilon$}
\Output{Processed Edge $E_{final}$}

\BlankLine
$G \leftarrow \text{BuildAdjacency}(E_{raw})$\;
$S_{avail} \leftarrow \text{Vertices}(G)$\;
$\tau \leftarrow \cos(\theta)$\;
$E_{\text{final}} \leftarrow \emptyset$\;

\BlankLine
\While{$S_{avail} \neq \emptyset$}{
    $v \leftarrow \text{PopArbitrary}(S_{avail})$\;
    $P \leftarrow [v]$\;
    \If{$G[v] \neq \emptyset$}{
        Append neighbor of $v$ to $P$ and update $G$\;
    }

    \Repeat{cannot extend}{
        \ForEach{$end \in \{\text{head, tail}\}$}{
            Let $\mathbf{d}$ be the direction vector at $end$\;
            Find neighbor $n \in G[P.\text{get}(end)]$ maximizing alignment with $\mathbf{d}$\;
            \If{alignment $> \tau$}{
                $P.\text{extend}(n)$ at $end$\;
                Remove edge to $n$ from $G$ and $S_{avail}$\;
            }
        }
    }

    \While{$|P| \ge k_{\min}$}{
        Find smallest index $i$ such that $\text{Dist}(P[i \dots \text{end}], \text{Line}(P[i], P[\text{end}])) < \epsilon$\;
        \eIf{segment found}{
            $E_{final}.\text{append}(\{P[i], P[\text{end}]\})$\;
            $P \leftarrow P[0 \dots i]$\;
        }{
            \textbf{break}\;
        }
    }
}
\Return{$E_{final}$}\;

\end{algorithm}

\section{Experiment Configurations}
\label{app:exp_config}

\textbf{Training Environment:} We train our policies using NVIDIA Isaac Sim and Isaac Lab \cite{mittal2025isaac} on an NVIDIA RTX 4090 GPU, parallelized with 4096 humanoid robot agents. We use the 29-DoF Unitree G1 humanoid robot for both simulation and physical deployment. The proprioceptive history length is $h=8$ steps, and the depth history uses $m=8$ strided frames with stride $\ell$.

\textbf{Hardware and Perception:} For real-world deployment, we use the factory-integrated Intel RealSense D435i depth camera (standard on Unitree G1, no hardware modification). Depth images are captured at 60~Hz at $480 \times 270$, downsampled to $64 \times 36$, and cropped to $32 \times 18$ as policy input.

\textbf{Onboard Deployment:} The policy runs at 50~Hz on the onboard NVIDIA Jetson Orin NX via \texttt{onnxruntime}. Depth acquisition and preprocessing run asynchronously at 60~Hz to minimize latency.

\textbf{Operator Guidance:} During hardware deployment, a human operator provides velocity commands through a remote controller, while the policy generates joint position targets from onboard observations. The position-based command generation procedure is used during training. Hardware operation therefore consists of perceptive locomotion under operator guidance. Policy changes follow a stop--stand--switch procedure: the robot stops, transitions to a standing policy, and then switches to the selected locomotion policy.

\textbf{Safety and Broader Impact:} Perceptive locomotion can extend humanoid operation to uneven outdoor terrain, but falls and unintended motion can endanger nearby people. Hardware testing should therefore remain under continuous human supervision, with an accessible emergency-stop mechanism. Tests in public spaces should use a clearly marked exclusion zone that keeps bystanders outside the robot's operating and potential fall area. Given the limited failure-recovery capability, the reported results do not establish safety for unsupervised operation around people.

\section{Simulation Evaluation Details}
\label{app:simulation_evaluation}

For the comparison and ablation study in \autoref{sec:comparison_ablation}, we evaluate each method with 1,000 parallel humanoid agents in simulation and report the success rate and mean reaching time in \autoref{tab:ablation_result}. A trial is considered successful if the robot reaches within 0.25~m of the target position within 15~s without triggering termination. The reaching time is computed for successful trials; entries marked as ``--'' indicate cases where no trial succeeds within the 15~s time limit, and the reaching time is therefore not reported.

All methods are tested on the same terrain categories in \autoref{tab:ablation_result}, including large box, ramp, small box, rough terrain, stair down, platform down, stair up, platform up, and gap. To make the command-generation ablation more discriminative, we use challenging evaluation terrains with the final geometry settings summarized in \autoref{tab:eval_terrain_settings}. These terrains expose the failure mode of the \textit{w/o position-based} baseline, which often refuses to climb the obstacle or turns around instead of reaching the specified target.

\begin{table}[H]
\centering
\small
\caption{Terrain geometry settings used for the simulation comparison and ablation study.}
\label{tab:eval_terrain_settings}
\renewcommand{\arraystretch}{1.15}
\setlength{\tabcolsep}{8pt}
\begin{tabular}{l p{0.66\linewidth}}
\toprule
\textbf{Terrain Type} & \textbf{Evaluation Geometry} \\
\midrule
Stair up/down & Step height $0.20$~m \\
Large box & Height $0.40$~m \\
Platform up/down & Platform height $0.40$~m \\
Ramp & Slope $0.40$~rad \\
Gap & Gap distance $0.60$~m \\
Rough terrain & Noise scale $0.05$ \\
Small box & Height $[0.25, 0.35]$~m; width and length randomly sampled from $[0.30, 0.50]$~m \\
\bottomrule
\end{tabular}
\end{table}

Because related perceptive locomotion methods do not provide directly reusable implementations or a shared benchmark, we use a best-effort reimplementation of HPL~\cite{zhuang2024humanoid} for the Unitree G1 in our training environment. The simulation comparison uses one generalist walking policy per method across all terrain categories.

\end{document}